\documentclass[conference]{IEEEtran}
\IEEEoverridecommandlockouts
\usepackage{cite}
\usepackage{amsmath,amssymb,amsfonts}
\usepackage{graphicx}
\usepackage{textcomp}
\usepackage{graphicx}
\usepackage{caption}
\usepackage{enumitem}
\usepackage[table]{xcolor}

\usepackage{algorithm}
\usepackage{amsthm}
\theoremstyle{plain}

\usepackage{algpseudocode}
\usepackage{url}
\usepackage[hidelinks]{hyperref}
\usepackage{multirow}
\usepackage{booktabs}
\usepackage{fancyhdr}
\def\BibTeX{{\rm B\kern-.05em{\sc i\kern-.025em b}\kern-.08em
    T\kern-.1667em\lower.7ex\hbox{E}\kern-.125emX}}
\begin{document}

\title{Residual-Tail-Informed Flow Matching for Time Series Anomaly Detection}

\author{
\IEEEauthorblockN{
Shyam Sundar Saravanan\IEEEauthorrefmark{1},
Tie Luo\IEEEauthorrefmark{2},
Sanjay Madria\IEEEauthorrefmark{1}
}
\IEEEauthorblockA{
\IEEEauthorrefmark{1}
Department of Computer Science,
Missouri University of Science and Technology,
Rolla, MO, USA\\
\{ssdmw, madrias\}@mst.edu
}
\IEEEauthorblockA{
\IEEEauthorrefmark{2}
Department of Electrical and Computer Engineering,
University of Kentucky,
Lexington, KY 40506, USA\\
t.luo@uky.edu
}
}

\maketitle

\fancypagestyle{firstpage}{
    \fancyhf{}
    \fancyfoot[C]{\copyright 2026 IEEE. Personal use of this material is permitted. Permission from IEEE must be obtained for other uses.}
    \renewcommand{\headrulewidth}{0pt}
}
\thispagestyle{firstpage}

\begin{abstract}

Time series anomaly detection (TSAD) models are often trained on nominal data using average-risk objectives, which emphasize dominant normal patterns. However, real time series can contain rare but valid nominal regimes that are underrepresented in the training distribution. We show that data-space flow matching can over-penalize such rare-normal windows and assign them anomaly scores comparable to those of true anomalies. To address this, we propose \emph{Residual-Tail-Informed Flow Matching} (RT-FM), an unsupervised framework that uses a temporal variational autoencoder to estimate residual latent tailness and converts it into bounded, warmup-annealed weights for training a data-space flow matching model. Experiments on seven TSAD benchmarks show that RT-FM improves rare-normal coverage while preserving strong anomaly discrimination, substantially reducing false alarms on such windows and resulting in improved aggregate detection performance. The code is available at: \url{https://github.com/trustAIoT/RT-FM_time-series-anomaly}

\end{abstract}


\section{Introduction}

Time series anomaly detection (TSAD) is commonly formulated as a one-class learning problem: a model is trained on nominal data and later used to score future windows according to deviation from learned normal behavior. This setting arises in industrial monitoring, cyber-physical systems, server telemetry, medical signals, and spacecraft data~\cite{hundman2018detecting,xu2022anomaly,su2019robust,tuli2022tranad}.

A key challenge is that nominal behavior is not homogeneous. Real time series can contain rare but valid operating regimes that are underrepresented during training. Conventional average-risk objectives can therefore fit dominant nominal behavior well while under-modeling infrequent nominal windows, leading to inflated anomaly scores. We use \emph{rare-normal} to denote a valid yet underrepresented nominal window. In general, since standard TSAD benchmarks do not provide rare-normal annotations, RT-FM uses {\em residual latent tailness} as an operational proxy for this concept.

\begin{figure*}[t]
    \centering
    \includegraphics[width=\textwidth]{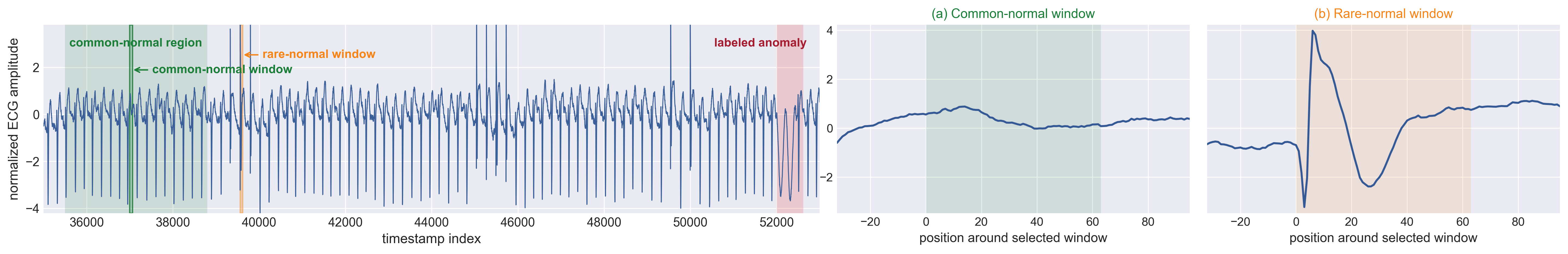}
    \caption{Motivating example from the UCR benchmark. The inset zooms into the selected rare-normal window, illustrating that nominal regions can contain heterogeneous behavior.}
    \label{fig:motivating-rare-normal}
\end{figure*}

Figure~\ref{fig:motivating-rare-normal} illustrates the motivating failure mode on a UCR sequence. The selected rare-normal window differs from the dominant local morphology while remaining outside the labeled anomaly interval. This example motivates improving coverage of underrepresented nominal structure without broadening the nominal model so aggressively that anomalous behavior is absorbed.

Flow matching (FM) learns a vector field transporting samples from a simple base distribution to the data distribution~\cite{lipman2023flow}. We propose \emph{Residual-Tail-Informed Flow Matching} (RT-FM), an unsupervised framework that uses a temporal VAE as a tail-discovery module while keeping the final detector in data space. The VAE estimates residual latent tailness; RT-FM removes the component explained by channel-wise amplitude and converts the resulting residual-tail score into bounded, warmup-annealed weights for training a data-space flow matching model. RT-FM uses the same training windows and flow-matching backbone as standard unweighted FM (\emph{Uniform FM}); the distinction lies in the training objective, where RT-FM redistributes influence to windows estimated to be underrepresented. Unlike detector-loss-based reweighting or post-hoc score-tail calibration ~\cite{lin2017focal,siffer2017anomaly}, RT-FM derives its weights from a detached nominal representation and applies them during flow-matching training.

Across seven univariate and multivariate benchmarks, RT-FM improves aggregate detection performance and reduces false positives at high recall. Controlled ablations isolate the benefit of residual-tail weighting from the underlying FM backbone. Our contributions are:
\begin{itemize}[leftmargin=*,nosep]
    \item We identify rare-normal over-penalization as a challenging failure mode of average-risk generative TSAD methods.
    \item We introduce RT-FM, which combines amplitude-residualized VAE tail scoring with bounded target-window weighting for data-space flow matching.
    \item We evaluate RT-FM on UCR~\cite{keogh2021dataset}, GECCO ~\cite{moritz_2018_3884398}, SWAN~\cite{2020NatSD}, MSL and SMAP, ~\cite{hundman2018detecting} PSM, and SMD~\cite{abdulaal2021practical,su2019robust} and provide controlled tail-signal and weighting-strength ablations.
\end{itemize}

\section{Related Work}

TSAD methods span recurrent, reconstruction, attention/Transformer, contrastive, and generative approaches~\cite{su2019robust,audibert2020usad,xu2022anomaly,tuli2022tranad,geiger2020tadgan}. Reconstruction-based detectors can reconstruct anomalies too well or infrequent nominal behavior poorly; attention-based models improve long-range dependency modeling but do not directly address imbalance between dominant and rare nominal regimes.

Generative TSAD methods use reconstruction, likelihood, or latent statistics for scoring. VAEs, normalizing flows, diffusion models, and flow matching provide complementary formulations~\cite{kingma2014auto,dinh2017density,papamakarios2021normalizing,ho2020denoising,lipman2023flow,tong2024improving}. RT-FM uses the VAE only to learn nominal latent geometry; anomaly scoring remains a data-space flow-matching procedure.

Tail-aware learning changes training emphasis through reweighting or risk-sensitive objectives, while EVT-style approaches typically model score tails at the decision stage~\cite{lin2017focal,rockafellar2000optimization,namkoong2017variance,siffer2017anomaly}.

\section{Methodology}

\begin{figure*}[t]
    \centering
    \includegraphics[width=\textwidth]{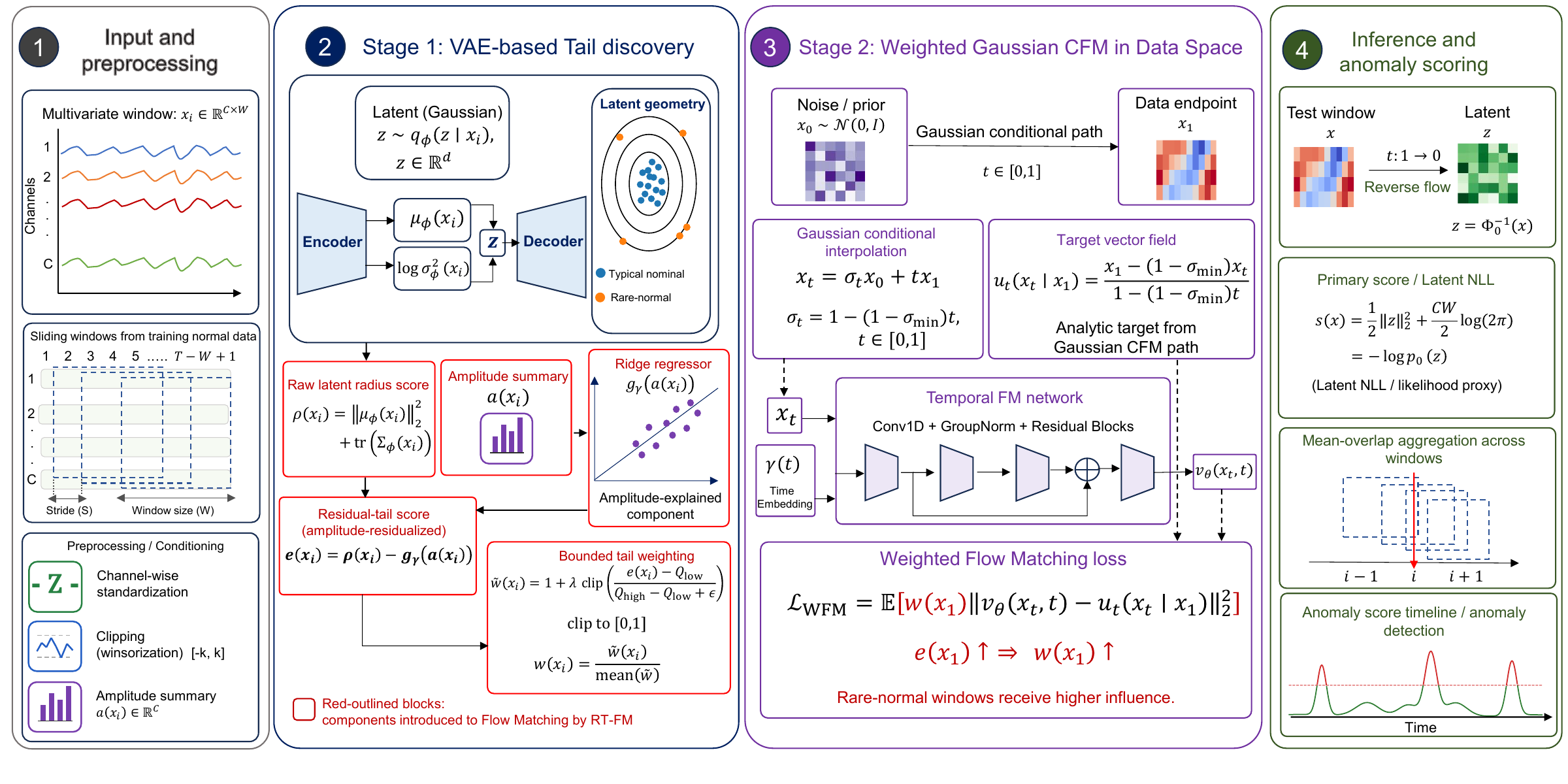}
    \caption{
    Overview of Residual-Tail-Informed Flow Matching (RT-FM). 
    Stage~1 trains a temporal VAE on nominal windows to estimate latent tailness.
    The posterior radius $\rho$ is residualized against channel-wise amplitude features
    to obtain a residual-tail score \(e\), which is converted into bounded
    target-window weights \(w(x_1)\). Stage~2 trains a data-space conditional flow
    matching model using these fixed target-window weights. At inference, test
    windows are mapped backward through the learned flow and scored using a
    Gaussian negative log-likelihood proxy, followed by mean-overlap aggregation
    to obtain timestamp-level anomaly scores.
    }
    \label{fig:architecture}
\end{figure*}

\subsection{Problem Setup and Window Construction}

For generality, consider multivariate time series. Let $Y=\{y_t\}_{t=1}^{T}\,,  y_t\in\mathbb{R}^{C},$ denote a time series with \(C\) channels and \(T\) timestamps. We construct overlapping windows of length \(W\) as
\[
x_i =
\big[y_i, y_{i+1}, \ldots, y_{i+W-1}\big]
\in \mathbb{R}^{C \times W}.
\]

Before windowing, each channel is standardized using statistics computed only
from the training sequence:
\[
\tilde{y}_{t,c}
=
\mathrm{clip}
\left(
\frac{y_{t,c}-\mu_c}{\max(\sigma_c,\epsilon)},
-\kappa,\kappa
\right),
\]
where \(\mu_c\) and \(\sigma_c\) are computed from the training data for channel
\(c\); the additional \(\epsilon\) is a scale floor and \(\kappa\) is the clipping threshold, which are introduced for numerical stability. During inference, this channel-wise transformation is identically applied to the test sequence to prevent data leakage.

\subsection{Residual-Tail Scoring with VAE}
\label{sec:residual-tail-scoring}

Rare-normality denotes valid but underrepresented nominal behavior. Since it is not directly observed, we use the residual-tail score \(e(x)\) as a proxy for it. To identify such windows, we train a temporal VAE on training windows treated as nominal. The encoder maps each multivariate window
\(x\in\mathbb{R}^{C\times W}\) to a Gaussian posterior
\[
q_\phi(z\mid x)
=
\mathcal{N}
\left(
\mu_\phi(x),
\operatorname{diag}(\sigma_\phi^2(x))
\right),
\]
where \(\phi\) denotes the encoder parameters, \(z\in\mathbb{R}^{d_z}\) is the latent representation with dimension \(d_z\), and
\(\mu_\phi(x),\sigma_\phi^2(x)\in\mathbb{R}^{d_z}\) are the posterior mean and variance vectors predicted by the encoder, respectively. The operator
\(\operatorname{diag}(\cdot)\) constructs a diagonal covariance matrix from the variance vector.

The VAE, with encoder parameters \(\phi\) and decoder parameters \(\psi\),
is optimized using a reconstruction term and a dimension-normalized KL
penalty. We define
\[
\mathcal{L}_{\mathrm{rec}}(x)
=
\frac{1}{CW}
\|x-\hat{x}\|_F^2
\]
\[
\mathcal{L}_{\mathrm{KL}}(x)
=
\frac{1}{d_z}
\sum_{j=1}^{d_z}
D_{\mathrm{KL}}
\left(
q_\phi(z_j\mid x)
\,\|\,\mathcal{N}(0,1)
\right).
\]
Let \(p_1\) denote the empirical distribution of training windows treated as nominal. The resulting VAE objective is
\[
\mathcal{L}_{\mathrm{VAE}}(\phi,\psi)
=
\mathbb{E}_{x\sim p_1}
\left[
\mathcal{L}_{\mathrm{rec}}(x)
+
\beta_s\mathcal{L}_{\mathrm{KL}}(x)
\right],
\]
where the KL coefficient is linearly warmed up as
\[
\beta_s
=
\beta_{\max}
\min
\left(
1,
\frac{s}{S_{\mathrm{KL}}}
\right),
\]
with \(\beta_{\max}=0.05\) and \(S_{\mathrm{KL}}=2500\) optimization steps.
Averaging the KL contribution across latent dimensions prevents its scale
from growing with \(d_z\), while the warmup reduces excessive early
regularization during representation learning. These fixed settings are used across datasets to avoid dataset-specific hyperparameter tuning while limiting early KL domination.

After training, we compute a radius-based posterior statistic for each
window, which is later used to estimate latent tailness:
\[
\rho(x)
=
\|\mu_\phi(x)\|_2^2
+
\sum_{j=1}^{d_z}\sigma_{\phi,j}^2(x).
\]
Since the raw latent radius \(\rho(x)\) can correlate with simple
amplitude variation, we remove its amplitude-explained component. Let
\(a(x)\in\mathbb{R}^{C}\) denote the vector of channel-wise standard
deviations within window \(x\). We fit a lightweight regressor
\(g_\gamma(a(x))\) from the channel-wise amplitude features to
\(\rho(x)\) using the nominal training set, and define the residual-tail
score as
\[
e(x)
=
\rho(x)-g_\gamma(a(x)).
\]
A large positive value of \(e(x)\) indicates a window that lies in a
peripheral region of the nominal latent geometry after accounting for
channel-wise amplitude.

Let \(Q_{\mathrm{low}}\) and \(Q_{\mathrm{high}}\) denote the empirical
\(0.90\)- and \(0.99\)-quantiles of \(\{e(x_i)\}_{i=1}^{N}\),
respectively. We define the residual-tail region as
\[
\mathcal{T}
=
\{x_i : e(x_i)\geq Q_{\mathrm{low}}\}.
\]
The residual-tail score is converted into a bounded target-window weight:
\[
\tilde{w}(x_i)
=
1
+
\lambda\,
\mathrm{clip}
\left(
\frac{e(x_i)-Q_{\mathrm{low}}}
{Q_{\mathrm{high}}-Q_{\mathrm{low}}+\epsilon},
0,1
\right),
\]
where \(\lambda\geq0\) controls the strength of residual-tail emphasis and
\(\epsilon>0\) prevents division by zero. The weights are then normalized
to have empirical mean one:
\[
w(x_i)
=
\frac{\tilde{w}(x_i)}
{\frac{1}{N}\sum_{j=1}^{N}\tilde{w}(x_j)}.
\]

{\bf Why a detached latent-tail signal?}
The VAE posterior radius is used as a compact geometry-based rarity signal while alternative methods have undesirable effects. A detector-loss signal would couple the weighting rule to the same model being optimized, making high-loss targets receive larger influence; this works against our goal of identifying underrepresented nominal structure independent of the detector. VAE reconstruction error can be dominated by decoder fidelity and local amplitude or waveform complexity, rather than by location in nominal representation space. A direct kNN density estimator is conceptually viable, but its behavior depends strongly on the representation, distance metric, and neighborhood size, and can be expensive on millions of overlapping windows. In contrast, RT-FM uses a frozen, low-dimensional posterior statistic and removes the component predictable from channel-wise amplitude. Sec.~\ref{ablation:boost-strength} verifies whether this residualization provides useful selectivity beyond the raw-radius and amplitude-only controls.

\subsection{Weighted Conditional Flow Matching}
\label{sec:conditional-flow-matching}
\label{sec:rtw-cfm}

After computing residual-tail weights, we train conditional flow matching (CFM) from a Gaussian base \(x_0\sim\mathcal{N}(0,I)\) to nominal target windows \(x_1\sim p_1\). With \(t\sim\mathcal U(0,1)\), we use the Gaussian conditional path
\begin{equation}
x_t=\sigma_t x_0+t x_1,\qquad
\sigma_t=1-(1-\sigma_{\min})t,
\end{equation}
which yields the analytic target vector field
\begin{equation}
u_t(x_t\mid x_1)=
\frac{x_1-(1-\sigma_{\min})x_t}{1-(1-\sigma_{\min})t}.
\end{equation}
Standard CFM trains \(v_\theta(x_t,t)\) to match \(u_t\):
\begin{equation}
\mathcal{L}_{FM}(\theta)=
\mathbb{E}\!\left[\|v_\theta(x_t,t)-u_t(x_t\mid x_1)\|_2^2\right].
\end{equation}
We refer to this standard unweighted objective as \emph{Uniform FM} throughout the experiments.
RT-FM changes only the target-window contribution using the fixed VAE-derived weight:
\begin{equation}
\mathcal{L}_{wFM}(\theta)=
\mathbb{E}\!\left[
w(x_1)\|v_\theta(x_t,t)-u_t(x_t\mid x_1)\|_2^2
\right].
\end{equation}
Since Uniform FM and RT-FM use the same architecture, training windows, and inference rule, their comparison isolates the effect of target-window reweighting.

\subsection{Model Architecture}
\label{sec:model-architecture}

Our method is organized into two architectural phases (see Fig.~\ref{fig:architecture}). Stage~1 trains a
temporal VAE that discovers rare but nominal windows, and Stage~2 trains a FM network using the VAE-induced target-window weights. The VAE architecture is summarized in Table~\ref{tab:vae_arch} and the FM
architecture is summarized in Table~\ref{tab:fm_arch}.

\paragraph{Stage 1: VAE-based tail discovery}
The first stage trains a lightweight temporal VAE on nominal training windows.
The VAE is not used as the final anomaly detector, but instead defines a latent
geometry over nominal windows. After training, the VAE is frozen and its
posterior statistics are used to compute the residual-tail score \(e(x)\) and target-window weight \(w(x)\).

The VAE uses shape-preserving one-dimensional convolutions over windows
\(x\in\mathbb{R}^{C\times W}\), GroupNorm, residual blocks, and \texttt{SiLU}
activations. The convolutional encoder captures local temporal motifs while
preserving the channel-by-time structure of the input. 
GroupNorm stabilizes training across datasets with
different channel counts and batch statistics.

\paragraph{Stage 2: Weighted CFM in data space}
The second stage trains a FM network using the fixed target-window
weights from Stage~1. This FM model learns the time-dependent vector field
\(v_\theta(x_t,t)\) over the space of preprocessed windows. It receives the
intermediate state \(x_t\) and timestep \(t\), and outputs
\(v_\theta(x_t,t)\in\mathbb{R}^{C\times W}\), a vector field with the same
shape as the intermediate state \(x_t\). Each window is represented as \(x\in\mathbb{R}^{C\times W}\),
where \(C\) is the number of channels and \(W\) is the window length.

The FM network serves as the primary nominal generative model, which uses timestep embedding, an input Conv1D projection, residual blocks, GroupNorm, SiLU activations and an
unconstrained Conv1D output projection. The output
layer has no activation, allowing the predicted vector field to take arbitrary
positive or negative values as required to match the target vector field
\(u_t(x_t\mid x_1)\).

\begin{table}[t]
\caption{Temporal VAE architecture and training hyperparameters.}
\centering
\resizebox{\columnwidth}{!}{%
\begin{tabular}{ccccc}
\toprule
Operation & Kernel & Stride & Units / Shape & Activation \\
\midrule
\textbf{Encoder} \\
Conv1D + GroupNorm & \(3\) & \(1\) & hidden size \(64\) & SiLU \\
ResidualBlock + GroupNorm (x3) & \(3\) & \(1\) & hidden size \(64\) & SiLU \\
Flatten & -- & -- & \(64W\) & -- \\
Fully Connected & -- & -- & \(d_z\) for \(\mu_\phi(x)\) & -- \\
Fully Connected & -- & -- & \(d_z\) for \(\log\sigma_\phi^2(x)\) & -- \\
\midrule
\textbf{Decoder} \\
Latent input & -- & -- & \(B \times d_z\) & -- \\
Fully Connected + reshape & -- & -- & \(B \times 64 \times W\) & -- \\
ResidualBlock + GroupNorm (x3) & \(3\) & \(1\) & hidden size \(64\) & SiLU \\
Conv1D output & \(3\) & \(1\) & \(B \times C \times W\) & -- \\
\midrule
\textbf{Hyperparameters} \\
Window size \(W\) & \(64\) & Latent dim. \(d_z\) & \(24\) & Hidden size \(64\) \\
Batch size & \(128\) & Iterations & \(30{,}000\) & Learning rate \(1\mathrm{e}{-3}\) \\
Max KL weight \(\beta_{\max}\) & \(0.05\) & Weight decay & \(1\mathrm{e}{-4}\) & Optimizer AdamW \\
\bottomrule
\end{tabular}%
}
\label{tab:vae_arch}
\end{table}

\begin{table}[t]
\caption{Flow matching vector-field network and training hyperparameters.}
\centering
\resizebox{\columnwidth}{!}{%
\begin{tabular}{ccccc}
\toprule
Operation & Kernel & Stride & Units / Shape & Activation \\
\midrule
\textbf{Main Network} \\
Input \(x_t\) & -- & -- & \(B \times C \times W\) & -- \\
Timestep embedding & -- & -- & \(d_{\mathrm{emb}}\) & SiLU \\
Input Conv1D & \(3\) & \(1\) & hidden size & -- \\
ResidualBlock + GroupNorm (x5) & \(3\) & \(1\) & hidden size & SiLU \\
Conv1D output projection & \(3\) & \(1\) & \(B \times C \times W\) & -- \\
\midrule
\textbf{Output} \\
Predicted vector field & -- & -- &
\(v_\theta(x_t,t)\in\mathbb{R}^{C\times W}\) & -- \\
\midrule
\textbf{Hyperparameters} \\
Window size \(W\) & \(64\) & Hidden size & \(128\) & $d_{\mathrm{emb}}$ $128$ \\
Step size & $1$ & Iterations & \(30{,}000\) & Learning rate \(1\mathrm{e}{-4}\) \\
Batch size & \(128\) & Optimizer & AdamW & Weight decay \(1\mathrm{e}{-4}\) \\
LR scheduler & StepLR & LR step & \(4{,}000\) & LR Gamma $0.9$ \\
\bottomrule
\end{tabular}%
}
\label{tab:fm_arch}
\end{table}

\subsection{Training Procedure}
\label{sec:train}

Training is carried out in the two phases of Sec.~\ref{sec:model-architecture}. Stage~1 trains and freezes the VAE; Stage~2 trains the FM vector field with nominal target windows, Gaussian base samples, and uniformly sampled timesteps. To stabilize optimization, target-window weights are introduced linearly:
\[
w_s(x)=1+\eta_s(w(x)-1),\qquad
\eta_s=\min\!\left(1,\frac{s}{S_{\mathrm{warm}}}\right),
\]
where \(S_{\mathrm{warm}}=0.2S\) and \(S\) is the total number of FM steps. Early optimization is therefore close to Uniform FM, while later optimization follows the residual-tail weighted objective.

\subsection{Inference and Anomaly Scoring}
\label{sec:scoring}

After training, the flow matching model defines a nominal transport between
the Gaussian base distribution and the reweighted nominal window distribution.
To score a test window \(x\in\mathbb{R}^{C\times W}\), we integrate the learned
vector field backward from \(t=1\) to \(t=0\):
\[
    z
    =
    x_0
    =
    x_1
    +
    \int_{1}^{0}
    v_\theta(x_t,t)\,dt
\]
If \(x\) is compatible with the learned nominal flow, its reverse image \(z\)
should lie in a high-density region of the Gaussian base distribution. We
therefore use a Gaussian negative log-likelihood proxy:
\[
    s(x)
    =
    \frac{1}{2}\|z\|_2^2
    +
    \frac{CW}{2}\log(2\pi).
\]
Larger values of \(s(x)\) indicate that the test window maps to a lower-density region of the base distribution and is therefore less consistent with the learned nominal dynamics.

Window-level scores are converted to timestamp-level scores using mean-overlap aggregation: each window score is first assigned to all timestamps covered by that window, and the timestamp score is then the average over all covering
windows:
\[
    s_t
    =
    \frac{1}{|\mathcal{I}(t)|}
    \sum_{i\in\mathcal{I}(t)} s(x_i),
    \qquad
    \mathcal{I}(t)=\{i: t\in[i,i+W-1]\}.
\]
The resulting timestamp-level score sequence is used for evaluation. For
threshold-free metrics such as AUROC, AP, and VUS-PR, thresholds are swept over
the score sequence. For thresholded metrics, we report the best F1 over all
thresholds and compute FP@95R using the threshold that achieves 95\% recall.

\section{Experimental Setup}
\label{sec:experiment-setup}

We evaluate RT-FM against state-of-the-art time series anomaly detection
baselines using the metrics detailed in Sec. \ref{sec:metrics} in both univariate and multivariate settings. For univariate
anomaly detection, we use the UCR benchmark. For multivariate anomaly
detection, we evaluate on GECCO, SWAN, MSL, SMAP, PSM, and SMD. Dataset statistics are reported in Table \ref{tab:dataset-stats}.

\begin{table}[h]
\centering
\caption{Statistics of the datasets used in our experiments. AR denotes anomaly ratio.}
\label{tab:dataset-stats}
\resizebox{\columnwidth}{!}{%
\begin{tabular}{lccccccc}
\toprule
\multirow{2}{*}{\textbf{Property}} &
\multicolumn{7}{c}{\textbf{Dataset}} \\
\cmidrule(lr){2-8}
& UCR & SMAP & MSL & GECCO & SMD & PSM & SWAN \\
\midrule
\# Series / entities & 250       & 55      & 27     & 1      & 28      & 1      & 1      \\
\# Dimensions ($d$) & 1         & 25      & 55     & 9      & 38      & 25     & 38     \\
\# Train size       & 5,302,449 & 140,825 & 58,317 & 69,260 & 708,405 & 132,481 & 60,000 \\
\# Test size        & 14,051,317 & 444,035 & 73,729 & 69,261 & 708,420 & 87,841 & 60,000 \\
AR (\%)             & 0.225     & 13.13   & 10.72  & 1.05   & 27.76   & 5.77   & 32.60  \\
\bottomrule
\end{tabular}%
}
\end{table}

\subsection{Datasets}
\label{sec:datasets}

UCR contains 250 independently curated univariate anomaly-detection series~\cite{keogh2021dataset}. Accordingly, Table~\ref{tab:dataset-stats} reports the number of evaluated series/entities rather than treating each series as a separate dataset. The benchmark spans heterogeneous domains and anomaly morphologies, making its aggregate results a broad univariate evaluation rather than repeated partitions of one underlying signal.

The multivariate benchmarks cover several operating regimes: SMAP and MSL~\cite{hundman2018detecting} contain NASA spacecraft telemetry; PSM and SMD ~\cite{abdulaal2021practical, su2019robust} contain server and infrastructure monitoring traces; GECCO ~\cite{moritz_2018_3884398} contains multivariate water-quality measurements; and SWAN~\cite{2020NatSD} contains space-weather monitoring signals. Their channel counts and anomaly ratios differ substantially (Table~\ref{tab:dataset-stats}), which provides a complementary test of RT-FM beyond the univariate UCR setting. These benchmarks annotate anomalies but do not independently identify rare-normal windows. Consequently, we use the residual-tail subset only for mechanism analysis, while overall detection metrics and all-nominal false-positive rates remain label-based evaluations.

Following the standard unsupervised benchmark protocol, training windows are treated as nominal when fitting both the VAE and the flow model. RT-FM therefore assumes that training contamination is limited rather than absent; the implications of stronger contamination are discussed in Sec.~\ref{sec:inference-limitations}.

\begin{table*}[t]
\centering
\caption{Aggregate UCR results over 250 series}
\label{tab:ucr_aggregate_results}
\resizebox{\textwidth}{!}{%
\begin{tabular}{|l|c|c|c|c|c|c|c|}
\hline
\textbf{Model} &
\textbf{F1} $\uparrow$ &
\textbf{Precision} $\uparrow$ &
\textbf{VUS-PR} $\uparrow$ &
\textbf{AUROC} $\uparrow$ &
\textbf{AP} $\uparrow$ &
\textbf{FP@95R} $\downarrow$ &
\textbf{Event F1} $\uparrow$ \\
\hline
TadGAN~\cite{geiger2020tadgan} & 0.131 & 0.092 & 0.332 & 0.641 & 0.118 & 0.617 & 0.154 \\
\hline
DENSE-AE~\cite{alnegheimish2022sintel} & 0.159 & 0.136 & 0.352 & 0.662 & 0.147 & 0.584 & 0.181 \\
\hline
LSTM-DT~\cite{alnegheimish2022sintel} & 0.320 & 0.289 & 0.442 & 0.731 & 0.306 & 0.462 & 0.342 \\
\hline
DONUT~\cite{donut18} & 0.351 & 0.325 & 0.428 & 0.726 & 0.337 & 0.439 & 0.371 \\
\hline
LSTM-AE~\cite{alnegheimish2022sintel} & 0.355 & 0.301 & 0.472 & 0.754 & 0.351 & 0.427 & 0.384 \\
\hline
MERLIN~\cite{nakamura2020merlin} & 0.414 & 0.402 & 0.503 & 0.781 & 0.419 & 0.386 & 0.492 \\
\hline
TSI-GAN~\cite{saravanan2023tsi} & 0.408 & 0.425 & 0.529 & 0.804 & 0.462 & 0.369 & 0.489 \\
\hline
DiffAD~\cite{xiao2023imputation} & 0.479 & 0.452 & 0.512 & 0.811 & 0.474 & 0.337 & 0.501 \\
\hline
TranAD~\cite{tuli2022tranad} & 0.512 & 0.470 & 0.561 & 0.829 & 0.586 & 0.306 & 0.532 \\
\hline
ImDiffusion~\cite{chen2023imdiffusion} & 0.548 & 0.530 & 0.579 & 0.801 & 0.527 & 0.298 & 0.521 \\
\hline
SGFM~\cite{he2024sgfm} & 0.529 & 0.515 & 0.612 & 0.861 & 0.584 & 0.357 & 0.603 \\
\hline
CARLA~\cite{darban2025carla} & 0.597 & 0.560 & 0.609 & 0.872 & 0.603 & 0.239 & 0.621 \\
\hline
Anomaly Transformer~\cite{xu2022anomaly} & 0.637 & 0.590 & 0.631 & 0.887 & 0.642 & 0.216 & 0.663 \\
\hline
DCDetector~\cite{yang2023dcdetector} & 0.672 & 0.602 & 0.651 & 0.899 & 0.675 & 0.201 & 0.692 \\
\hline
\rowcolor{green!15}
RT-FM & \textbf{0.723} & \textbf{0.692} & \textbf{0.674} & \textbf{0.914} & \textbf{0.738} & \textbf{0.184} & \textbf{0.761} \\
\hline
\end{tabular}
}
\end{table*}

\begin{table*}[h]
\centering
\caption{Multivariate TSAD results}
\label{tab:multivariate-results}
\resizebox{\textwidth}{!}{%
\begin{tabular}{lcccccccccccc}
\toprule
\multirow{2}{*}{Method} 
& \multicolumn{2}{c}{GECCO} 
& \multicolumn{2}{c}{SWAN} 
& \multicolumn{2}{c}{MSL} 
& \multicolumn{2}{c}{SMAP} 
& \multicolumn{2}{c}{PSM} 
& \multicolumn{2}{c}{SMD} \\
\cmidrule(lr){2-3}
\cmidrule(lr){4-5}
\cmidrule(lr){6-7}
\cmidrule(lr){8-9}
\cmidrule(lr){10-11}
\cmidrule(lr){12-13}
& VUS-PR $\uparrow$ & F1 $\uparrow$
& VUS-PR $\uparrow$ & F1 $\uparrow$
& VUS-PR $\uparrow$ & F1 $\uparrow$
& VUS-PR $\uparrow$ & F1 $\uparrow$
& VUS-PR $\uparrow$ & F1 $\uparrow$
& VUS-PR $\uparrow$ & F1 $\uparrow$ \\
\midrule
OmniAnomaly         & 0.612 & 0.661 & 0.746 & 0.790 & 0.628 & 0.678 & 0.578 & 0.635 & 0.703 & 0.742 & 0.677 & 0.725 \\
LSTM-AE             & 0.637 & 0.684 & 0.768 & 0.813 & 0.650 & 0.700 & 0.606 & 0.662 & 0.724 & 0.761 & 0.693 & 0.740 \\
DCDetector          & 0.681 & 0.728 & 0.805 & 0.846 & 0.698 & 0.745 & 0.748 & 0.701 & 0.756 & 0.789 & 0.733 & 0.780 \\
USAD                & 0.701 & 0.746 & 0.825 & 0.859 & 0.720 & 0.761 & 0.678 & 0.725 & 0.772 & 0.804 & 0.748 & 0.792 \\
TranAD              & 0.724 & 0.721 & 0.839 & 0.878 & 0.719 & 0.729 & 0.698 & 0.748 & 0.791 & 0.821 & 0.784 & 0.808 \\
CARLA               & 0.719 & 0.781 & 0.852 & 0.889 & 0.763 & 0.804 & 0.712 & 0.761 & 0.804 & 0.833 & 0.726 & 0.801 \\
SGFM                & 0.751 & 0.792 & 0.827 & 0.871 & 0.769 & 0.808 & 0.710 & 0.767 & 0.813 & 0.839 & 0.782 & 0.822 \\
DiffAD              & 0.766 & 0.806 & 0.862 & 0.899 & 0.782 & 0.821 & 0.738 & 0.782 & 0.826 & 0.852 & 0.759 & 0.821 \\
ImDiffusion         & 0.779 & 0.818 & 0.864 & 0.901 & 0.793 & 0.832 & 0.750 & 0.794 & 0.833 & 0.861 & 0.762 & 0.835 \\
Anomaly Transformer & 0.801 & 0.841 & 0.884 & 0.918 & 0.819 & 0.857 & 0.772 & 0.812 & 0.856 & 0.879 & \textbf{0.883} & \textbf{0.860} \\
\rowcolor{green!12}
RT-FM               & \textbf{0.893} & \textbf{0.872} & \textbf{0.966} & \textbf{0.935} & \textbf{0.896} & \textbf{0.872} & \textbf{0.858} & \textbf{0.843} & \textbf{0.939} & \textbf{0.901} & 0.856 & 0.842 \\
\bottomrule
\end{tabular}%
}
\end{table*}

\subsection{Baselines and Evaluation}
\label{sec:baselines}
\label{sec:metrics}

We compare RT-FM with reconstruction, Transformer, diffusion, contrastive, and flow-based baselines: LSTM-AE, USAD~\cite{audibert2020usad}, OmniAnomaly~\cite{su2019robust}, DONUT~\cite{donut18}, TadGAN~\cite{geiger2020tadgan}, TSI-GAN~\cite{saravanan2023tsi}, TranAD~\cite{tuli2022tranad}, Anomaly Transformer~\cite{xu2022anomaly}, DiffAD~\cite{xiao2023imputation}, ImDiffusion~\cite{chen2023imdiffusion}, DCDetector~\cite{yang2023dcdetector}, CARLA~\cite{darban2025carla}, and SGFM~\cite{he2024sgfm}. This set covers both widely used TSAD families and a recent flow-matching baseline, allowing the experiments to separate the value of flow-based generative modeling from the additional effect of residual-tail weighting.

Official implementations are used whenever available. SGFM, whose official implementation is not publicly available, is reimplemented by us based on its paper. Hyperparameters follow recommended values whenever specified; otherwise, they are selected from a small training-only grid based on convergence and training-loss behavior. Baseline predictions are evaluated using the same timestamp-level metrics and aggregation protocol as RT-FM, so the reported comparisons do not mix method-specific thresholding conventions.

We report AUROC and AP as threshold-free ranking metrics and VUS-PR to incorporate temporal tolerance around anomaly intervals. For thresholded evaluation, best timestamp-level F1 summarizes the strongest precision--recall tradeoff, event F1 evaluates overlap with contiguous anomaly intervals, and FP@95R measures false-alarm burden at a fixed high-recall operating point. Window scores are converted to timestamps by the mean-overlap aggregation of Sec.~\ref{sec:scoring}. Because rare-normal over-penalization manifests primarily as false alarms on nominal behavior, FP@95R is particularly useful here, while AUROC/AP/VUS-PR ensure that improvements are not an artifact of one selected threshold.

\section{Experimental Results}

\begin{table}[t]
\centering
\caption{Aggregate multivariate TSAD results across GECCO, SWAN, MSL, SMAP, PSM, and SMD.}
\label{tab:multivariate-agg-results}
\resizebox{\columnwidth}{!}{%
\begin{tabular}{lcccc}
\toprule
Method & AUROC $\uparrow$ & AP $\uparrow$ & FP@95R $\downarrow$ & Event F1 $\uparrow$ \\
\midrule
OmniAnomaly         & 0.781 & 0.521 & 0.774 & 0.705 \\
LSTM-AE             & 0.796 & 0.548 & 0.751 & 0.727 \\
DCDetector          & 0.813 & 0.592 & 0.704 & 0.782 \\
USAD                & 0.831 & 0.611 & 0.621 & 0.798 \\
TranAD              & 0.816 & 0.637 & 0.646 & 0.821 \\
CARLA               & 0.828 & 0.612 & 0.623 & 0.835 \\
SGFM                & 0.862 & 0.621 & 0.611 & 0.812 \\
DiffAD              & 0.871 & 0.614 & 0.592 & 0.854 \\
ImDiffusion         & 0.876 & 0.697 & 0.579 & 0.860 \\
Anomaly Transformer & 0.889 & 0.724 & 0.548 & \textbf{0.878} \\
\rowcolor{green!12}
RT-FM & \textbf{0.903} & \textbf{0.771} & \textbf{0.522} & 0.840 \\
\bottomrule
\end{tabular}%
}
\end{table}

\begin{figure*}[t]
    \centering
    \includegraphics[width=\textwidth]{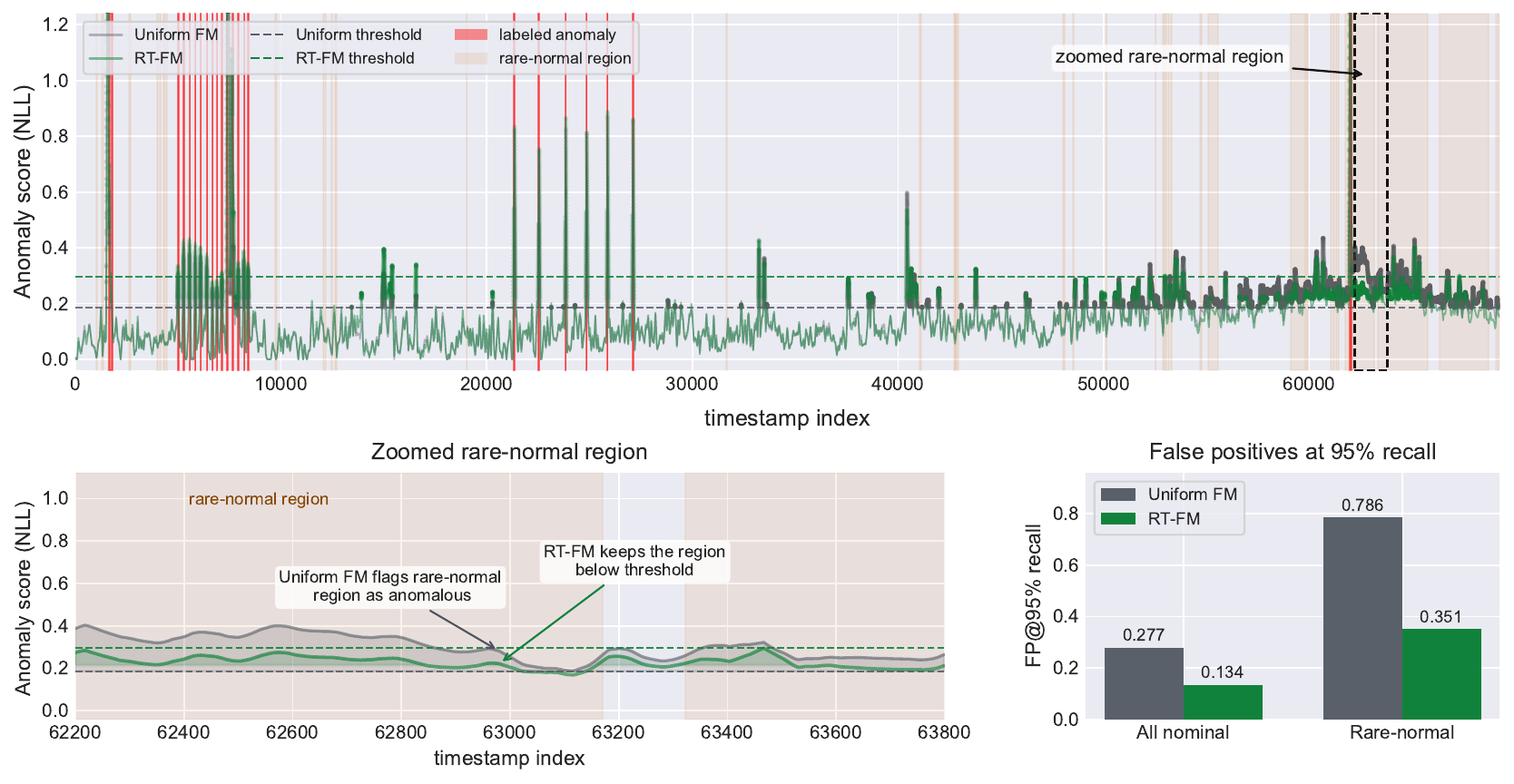}
    \caption{Residual-tail nominal false-positive analysis on GECCO. Uniform FM assigns elevated NLL scores to highlighted nominal regions outside labeled anomalies, while RT-FM lowers these scores. FP@95R is reported over all nominal timestamps and over the operational residual-tail nominal subset.}
    \label{fig:rare-normal-fp-analysis}
\end{figure*}

\subsection{Main Results}

Table~\ref{tab:ucr_aggregate_results} shows that RT-FM achieves the best reported UCR performance across all metrics, including F1 \(0.723\), precision \(0.692\), VUS-PR \(0.674\), AUROC \(0.914\), AP \(0.738\), and FP@95R \(0.184\). Relative to DCDetector, F1 and AP improve by \(7.6\%\) and \(9.3\%\), precision improves by \(15.0\%\), and FP@95R decreases by \(8.5\%\). These gains show that the improvement is not limited to one thresholded metric, but appears in both ranking quality and false-positive behavior at high recall.

On the six multivariate benchmarks (Table~\ref{tab:multivariate-results}), RT-FM obtains the best VUS-PR and F1 on five datasets. On GECCO, VUS-PR improves from 0.801 to 0.893 and F1 from 0.841 to 0.872 relative to Anomaly Transformer; on PSM, the corresponding gains are 0.856 to 0.939 and 0.879 to 0.901. SMD is the exception, where Anomaly Transformer obtains 0.883/0.860 versus RT-FM's 0.856/0.842. Aggregate AUROC/AP improve from 0.889/0.724 for Anomaly Transformer to 0.903/0.771 for RT-FM, while FP@95R decreases from 0.548 to 0.522 (Table~\ref{tab:multivariate-agg-results}). The controlled Uniform-FM comparison in Sec.~\ref{ablation:boost-strength} separates these gains from the strength of the underlying flow-matching backbone.

\subsection{Ablation: Tail Signal and Boost Strength}
\label{ablation:boost-strength}

To isolate residual-tail weighting, the controlled variants share the same window construction, vector-field backbone, optimizer, and inference procedure. Uniform FM, defined in Sec.~\ref{sec:conditional-flow-matching}, measures the contribution of the FM backbone before any tail weighting is introduced. \textbf{Amplitude FM} weights target windows using channel-wise amplitude summaries. \textbf{Raw-radius FM} uses the VAE posterior radius directly as the tail signal. \textbf{Residual-tail FM} uses the proposed amplitude-residualized latent tail score. These variants differ only in the target-window weighting strategy, allowing us to assess whether residualization provides a more selective mechanism for emphasizing underrepresented nominal structure.

Table~\ref{tab:multivariate-overall} reports the controlled ablation results. Amplitude weighting performs below Uniform FM across the reported metrics, whereas raw-radius weighting improves progressively as the boost strength increases from \(\lambda=1\) to \(\lambda=2\). Residual-tail weighting provides a stronger improvement, with \(\lambda=2\) increasing AP from 0.706 to 0.738 and VUS-PR from 0.812 to 0.833 relative to Uniform FM, while reducing FP@95R from 0.572 to 0.526. Compared with Raw-radius FM at the same \(\lambda=2\), Residual-tail FM improves AP from 0.721 to 0.738, VUS-PR from 0.821 to 0.833, and FP@95R from 0.552 to 0.526, supporting the benefit of removing the amplitude-explained component from the latent tail signal.

This controlled comparison also clarifies how much of the final result comes from the FM backbone itself. Uniform FM is already competitive, confirming that the underlying data-space flow model is strong; residual-tail weighting then adds a consistent improvement in AUROC, AP, VUS-PR, and FP@95R. The contribution is therefore not that flow matching alone solves TSAD, but that the proposed nominal-coverage mechanism improves a fixed flow-matching detector without changing its architecture or inference rule.

We further vary the residual-tail boost strength over
\(\lambda\in\{1,2,4,5,6,10,20,50,100\}\). Increasing \(\lambda\) yields substantial gains in ranking-based detection performance up to a broad optimum at moderate-to-strong weighting. AUROC reaches its maximum of 0.905 at \(\lambda=10\), while AP and VUS-PR peak at 0.771 and 0.902, respectively, at \(\lambda=20\). The lowest FP@95R is obtained at \(\lambda=6\) with 0.497, with \(\lambda=10\) remaining nearly identical at 0.498. At larger values, performance begins to saturate and eventually decline: AP and VUS-PR remain high at \(\lambda=50\) but decrease to 0.764 and 0.883 at \(\lambda=100\). These results indicate that RT-FM is effective across a broad range of boost strengths, with the strongest overall ranking performance concentrated around \(\lambda=20\). At \(\lambda=20\), RT-FM attains its best AP and VUS-PR but yields lower Event F1 than Uniform FM (0.840 vs. 0.861), showing that the optimal boost strength depends on the evaluation metric.

\begin{table}[h]
\centering
\caption{Aggregate multivariate anomaly-detection performance using timestamp-level NLL scores.}
\label{tab:multivariate-overall}
\resizebox{\columnwidth}{!}{%
\begin{tabular}{lccccc}
\toprule
Method & AUROC $\uparrow$ & AP $\uparrow$ & VUS-PR $\uparrow$ &
FP@95R $\downarrow$ & Event F1 $\uparrow$ \\
\midrule
Uniform FM                         & 0.881 & 0.706 & 0.812 & 0.572 & 0.861 \\
Amplitude FM                       & 0.875 & 0.697 & 0.798 & 0.591 & 0.856 \\
Raw-radius FM ($\lambda=1$)        & 0.884 & 0.714 & 0.817 & 0.561 & 0.867 \\
Raw-radius FM ($\lambda=2$)        & 0.887 & 0.721 & 0.821 & 0.552 & 0.872 \\
Residual-tail FM ($\lambda=1$)     & 0.888 & 0.724 & 0.823 & 0.548 & 0.874 \\
Residual-tail FM ($\lambda=2$)     & 0.894 & 0.738 & 0.833 & 0.526 & 0.883 \\
Residual-tail FM ($\lambda=4$)     & 0.902 & 0.751 & 0.844 & 0.503 & \textbf{0.894} \\
Residual-tail FM ($\lambda=5$)     & 0.900 & 0.751 & 0.848 & 0.510 & 0.760 \\
Residual-tail FM ($\lambda=6$)     & 0.902 & 0.756 & 0.862 & \textbf{0.497} & 0.762 \\
Residual-tail FM ($\lambda=10$)    & \textbf{0.905} & 0.768 & 0.895 & 0.498 & 0.771 \\
Residual-tail FM ($\lambda=20$)    & 0.903 & \textbf{0.771} & \textbf{0.902} & 0.522 & 0.840 \\
Residual-tail FM ($\lambda=50$)    & 0.903 & 0.770 & 0.898 & 0.530 & 0.832 \\
Residual-tail FM ($\lambda=100$)   & 0.897 & 0.764 & 0.883 & 0.531 & 0.811 \\
\bottomrule
\end{tabular}%
}
\end{table}

\subsection{When Does Residual-Tail Weighting Help?}
RT-FM is most useful when a strong Uniform-FM backbone still assigns elevated scores to rare-normal windows. GECCO and PSM illustrate this regime, while the controlled ablation shows that amplitude weighting does not reproduce the gain and raw-radius weighting is weaker without amplitude residualization. RT-FM is not universally superior on every operating metric: SMD remains a counterexample, and \(\lambda=20\) trades Event F1 for stronger aggregate AP/VUS-PR. This tradeoff depends on how frequently underrepresented nominal windows occur and on the selected evaluation metric.

\subsection{\texorpdfstring{Residual-Tail Nominal False-Positive Analysis}{Residual-Tail Nominal False-Positive Analysis}}
\label{sec:rare_normal_fp}

At 95\% recall, RT-FM reduces false positives over \emph{all} nominal GECCO timestamps from 0.277 to 0.134; this comparison is independent of residual-tail subset selection. Within the operational residual-tail nominal subset, FP@95R decreases from 0.786 to 0.351 while the labeled anomaly response remains strong.

\subsection{Inference Efficiency and Limitations}
\label{sec:inference-limitations}

\begin{table}[h]
\centering
\caption{Inference time per window for RT-FM scoring.}
\label{tab:inference-time}
\resizebox{\columnwidth}{!}{%
\begin{tabular}{lccc}
\toprule
Input & Batch size & FM forward / window & NLL scoring / window \\
\midrule
Univariate (\(C=1\)) & 1   & 1.664 ms & 30.056 ms \\
Univariate (\(C=1\)) & 32  & 0.053 ms & 0.970 ms \\
Univariate (\(C=1\)) & 256 & 0.007 ms & 0.124 ms \\
\midrule
Multivariate (\(C=55\)) & 1   & 1.674 ms & 30.189 ms \\
Multivariate (\(C=55\)) & 32  & 0.053 ms & 0.951 ms \\
Multivariate (\(C=55\)) & 256 & 0.007 ms & 0.127 ms \\
\bottomrule
\end{tabular}}
\end{table}

Table~\ref{tab:inference-time} separates a single FM-network evaluation from full NLL scoring, which requires reverse ODE integration. Batching reduces univariate NLL cost from 30.056 ms/window at batch size 1 to 0.970 ms at batch size 32 and 0.124 ms at batch size 256; the multivariate case follows nearly the same pattern. The similar univariate and multivariate timings indicate that, under this architecture, inference cost is dominated primarily by ODE-solver and GPU-launch overhead rather than channel count.

These timings indicate that RT-FM is practical for batched offline scoring and high-throughput monitoring, although strict low-latency streaming remains limited by reverse ODE integration and may benefit from fewer solver steps, solver distillation, or approximate scoring. RT-FM also assumes training data are predominantly nominal. Bounded weights and warmup limit the influence of individual tail windows, but they do not guarantee robustness when contamination is substantial; contamination-robust tail discovery is therefore a meaningful extension.

\section{Conclusion and Future Work}
\label{sec:conclusion}

We presented RT-FM, a residual-tail-informed flow-matching framework for time series anomaly detection. RT-FM uses a temporal VAE to characterize nominal latent geometry, residualizes latent tailness against channel-wise amplitude, and converts the resulting scores into bounded target-window weights for data-space conditional flow matching. Across univariate and multivariate benchmarks, RT-FM improves aggregate detection performance and reduces false positives at high recall. Controlled ablations further show that amplitude-only weighting is an unreliable proxy for underrepresented nominal structure, while residual-tail weighting is more selective than raw latent-radius weighting.

Future work will study contamination-robust weighting, long-range tail discovery, and faster streaming inference.

\bibliographystyle{IEEEtran}
\bibliography{bigdata}

\end{document}